\documentclass[final]{l4dc2025}

\usepackage{physics}
\usepackage{booktabs}
\usepackage{enumerate}
\usepackage{caption}
\usepackage{subcaption}

\usepackage[bottom]{footmisc}

\title{Orthogonal projection-based regularization for efficient model augmentation}
\usepackage{times}

\coltauthor{%
 \Name{Bendegúz M. Györök}{\normalfont\textsuperscript{1}} \Email{gyorokbende@sztaki.hun-ren.hu}\\
 \Name{Jan H. Hoekstra}{\normalfont\textsuperscript{2}}
 \Email{j.h.hoekstra@tue.nl}\\
  \Name{Johan Kon}{\normalfont\textsuperscript{3}}
 \Email{j.j.kon@tue.nl}\\
 \Name{Tamás Péni}{\normalfont\textsuperscript{1,4}} \Email{peni@sztaki.hun-ren.hu}\\
 \Name{Maarten Schoukens}{\normalfont\textsuperscript{2}}
 \Email{m.schoukens@tue.nl}\\
 \Name{Roland Tóth}{\normalfont\textsuperscript{1,2}} \Email{r.toth@tue.nl}\\
 \addr{\normalfont\textsuperscript{1}}Systems and Control Lab, HUN-REN Institute for Computer Science and Control, Budapest, Hungary\\
{\normalfont\textsuperscript{2}}Control Systems Group, Eindhoven University of Technology, The Netherlands\\
{\normalfont\textsuperscript{3}}Control Systems Technology Group, Eindhoven University of Technology, The Netherlands\\
{\normalfont\textsuperscript{4}}National Laboratory for Health Security, Budapest, Hungary\vspace{-18pt}}

\begin{document}

\maketitle
\vspace{-12pt}
\begin{abstract}%
Deep-learning-based nonlinear system identification has shown the ability to produce reliable and highly accurate models in practice. However, these black-box models lack physical interpretability, and a considerable part of the learning effort is often spent on capturing already expected/known behavior of the system, that can be accurately described by first-principles laws of physics. A potential solution is to directly integrate such prior physical knowledge into the model structure, combining the strengths of physics-based modeling and deep-learning-based identification. The most common approach is to use an additive model augmentation structure, where the physics-based and the {\em machine-learning} (ML) components are connected in parallel, i.e., additively. However, such models are overparametrized, training them is challenging, potentially causing the physics-based part to lose interpretability. To overcome this challenge, this paper proposes an orthogonal projection-based regularization technique to enhance parameter learning and even model accuracy in learning-based augmentation of nonlinear baseline models.
\end{abstract}

\vspace{-12pt}
\section{Introduction}\label{sec:intro}
\vspace{-4pt}
Continuously rising performance requirements and the growing complexity of systems resulted in high demand for nonlinear models that can accurately describe complex behavior. For example, reliable path planning and motion control of autonomous vehicles require accurate dynamic models. {\em First-principle} (FP) models for these systems can usually be obtained from physical knowledge and practical engineering insight. While FP modeling approaches offer models that can describe the main dynamic behavior of systems on a wide operating range, they often provide approximative descriptions of complex dynamical aspects such as aerodynamic forces, tire dynamics, or friction characteristics, which can become dominant under high-performance operation of the system.

As an alternative approach, data-driven nonlinear modeling methods (identification) have been developed, capable of providing highly accurate and consistent models of the system behavior. In particular, recent approaches utilizing {\em state-space} (SS) models based on deep {\em artificial neural networks} (ANNs) have shown exceptional results \citep{masti_learning_2021}.
However, the use of such models in practical applications, such as trajectory planning and control raised some significant concerns, since they lack physical interpretation \citep{ljung_perspectives_2010}. Furthermore, such ANN-based black-box models typically extrapolate inaccurately from training data and are more prone to overfitting when trained on small datasets. Another drawback of black-box identification is that often a considerable time of the learning effort is spent on capturing already expected behavior due to first-principles-based understanding of some aspects of the system.

To address these challenges, \emph{physics-informed neural networks} (PINNs) and \emph{physics-guided neural networks} (PGNNs) were introduced in \cite{raissi_physics-informed_2019}, and \cite{daw_physics-guided_2022}, respectively. Both branches of model learning apply an additional term in the cost function, penalizing when the ANN predictions deviate from priori-selected physical laws during model training \citep{karpatne_theory-guided_2017}. A more traditional approach for combining physics-based knowledge and data-driven identification techniques is (light) grey-box modeling. A wide range of methods has been developed and applied successfully in practice for years \citep{bohlin_practical_2006}, mainly for \emph{linear time invariant} (LTI) system representations. Light grey-box models utilize physics-based system descriptions and data-driven parameter estimation \citep{schoukens_nonlinear_2019}.

Model augmentation, i.e., hybrid modeling, is a promising direction of utilizing physical knowledge in nonlinear system identification \citep{schon_multi-objective_2022}. By combining FP models with flexible learning components, (\textit{i}) faster convergence and (\textit{ii}) better accuracy can be achieved compared to black-box learning methods while producing (\textit{iii}) physically interpretable models \citep{djeumou_neural_2022}. Furthermore, existing approaches can be incorporated into the model augmentation framework, e.g., physics-based loss functions can be utilized during model optimization \citep{daw_physics-guided_2022} and physical parameters can be co-estimated with the parameters of the learning component \citep{psichogios_hybrid_1992}.

This paper investigates learning for hybrid model augmentation when the model is composed of an FP model with an additive neural network part, and the objective is to jointly estimate the ANN parameters and the physical parameters based on measured data from the system. This formulation can produce models with properties (\textit{i}-\textit{iii}), however simultaneous tuning of the learning-based and physical parameters causes overparametrization and results in "competing" submodels. Instead of only compensating for the unmodeled terms, the ML part can learn some of the known dynamics of the FP model as well. This can generate unrealistic parameters in the FP model, resulting in less interpretable models and even hurting extrapolation properties. Recent studies have addressed similar challenges in the context of PGNN-based feedforward control methods  \citep{bolderman_feedforward_2022, bolderman_physics-guided_2024}, yet, this issue remains unsolved for nonlinear system identification. Orthogonalization is an attractive approach that we now generalize for nonlinear model learning, inspired by \cite{kon_physics-guided_2022}. With this modification, the learning component is penalized for learning the known dynamics of the FP model, by forcing a particular form of orthogonality between the ML- and physics-based layers. The original approach is limited to FP models that are linear in their parameters; therefore, a generalized version of the method is also presented in this paper.

The main contributions of this work can be summarized as:\vspace{-6pt}
\begin{enumerate}[C1]
    \item Generalization of an orthogonal projection-based regularization method \citep{kon_physics-guided_2022} for FP models that are nonlinear in the parameters, and integrating the approach to learning-based model augmentation.\vspace{-6pt}
    \item Efficient initialization and normalization schemes for the additive model augmentation.
\end{enumerate}\vspace{-4pt}

The remainder of the paper is organized as follows: Sect.~\ref{sec:additive_augm} introduces the identification problem and the additive model augmentation structure. The applied cost function, normalization, and initialization schemes are also discussed. Then, Sect.~\ref{sec:orthogonalization} gives the orthogonal projection-based regularization term for FP models that are linear in the parameters, which is generalized in Sect.~\ref{sec:orth_generalization} to FP models that can be nonlinear in the parameters. Sect.~\ref{sec:f1tenth_ident} provides a simulation study to demonstrate the effectiveness of the methodology. Finally, in Sect.~\ref{sec:conclusion}, the conclusions on the presented work are drawn.
\vspace{-6pt}
\section{Additive model augmentation}\label{sec:additive_augm}
\vspace{-4pt}
\subsection{Problem formulation}
\vspace{-2pt}
The dynamics of the data-generating system are considered to be defined by a \emph{discrete-time} (DT) nonlinear SS representation:
\begin{subequations}
\label{eq:DT-SS}
\begin{align}
    x_{k+1}&=f(x_k, u_k),\\
    y_k&=h(x_k, u_k) + e_k,\label{eq:DT-y}
\end{align}
\end{subequations}
where $k\in\mathbb{Z}$ is the discrete time index, $x_k\in\mathbb{R}^{n_x}$ is the state, $u_k\in\mathbb{R}^{n_u}$ is the control input, $y_k\in\mathbb{R}^{n_y}$ is the measured output, $f:\mathbb{R}^{n_x} \times \mathbb{R}^{n_u} \rightarrow \mathbb{R}^{n_x}$ is the DT state transition function, $h:\mathbb{R}^{n_x} \times \mathbb{R}^{n_u} \rightarrow \mathbb{R}^{n_y}$ is the output map, and $e_k\in\mathbb{R}^{n_y}$ is an i.i.d. white noise process with finite variance, representing measurement noise.

The exact dynamics of \eqref{eq:DT-SS} are not known, but we assume that based on prior knowledge, a physics-based approximative model (baseline model) is available in the form of
\begin{subequations}
\label{eqs:fp_model}
\begin{align}
    \hat{x}_{k+1}&=f_\theta(\hat{x}_k,u_k),\\
    \hat{y}_k&=h_\theta(\hat{x}_k, u_k),
\end{align}
\end{subequations}
where $\hat{x}_k\in\mathbb{R}^{n_x}$ is the model state, $\hat{y}_k\in\mathbb{R}^{n_y}$ is the model output, $f_\theta:\mathbb{R}^{n_x} \times \mathbb{R}^{n_u} \rightarrow \mathbb{R}^{n_x}$ is the FP state transition function, and $h_\theta:\mathbb{R}^{n_x} \times \mathbb{R}^{n_u} \rightarrow \mathbb{R}^{n_y}$ is the output function, that both depend on physical parameters $\theta\in\mathbb{R}^{n_\theta}$. 
Nominal parameter values $\theta_0$ can often be obtained directly from manufacturer datasheets, as they typically provide specifications and coefficient values for many subcomponents. In cases where such information is unavailable, prior coefficient estimation may be required to determine suitable nominal values.

As discussed in Sect.~\ref{sec:intro}, even by optimizing the parameters of FP models, they can often only provide an approximative representation of the true underlying dynamics of the system, hence we can use deep-learning-based model augmentation to enhance their modeling capabilities. One of the most straightforward ways to augment the baseline first-principle dynamics is to introduce an additive term into the state transition \citep{sun_comprehensive_2020, sohlberg_grey_2008}.
The FP output function $h_\theta$ is assumed to contain simple relations (it can even be an identity map), thus, the aim is to identify the system dynamics in the form of
\begin{subequations}\label{eqs:additive_struct}
\begin{align}
    \hat{x}_{k+1} &= f_\theta(\hat{x}_k,u_k) + f^\mathrm{ANN}_\eta(\hat{x}_k, u_k),\\
    \hat{y}_k &= h_\theta(\hat{x}_k, u_k),
\end{align}
\end{subequations}
where $f^\mathrm{ANN}$ here represents a fully connected, feedforward neural network with $\eta\in\mathbb{R}^{n_\eta}$ collecting its parameters, but can be replaced by any function approximator without loss of generality.


Although nominal values are typically available, still certain physical parameters are often only approximately known. Hence, to get the best prediction performance, the first-principle model parameters are estimated simultaneously with the ANN parameters. 
The goal is to find the underlying physical parameters $\theta_\ast$ that resemble the actual system properties as closely as possible while finding $\eta_\ast$ parameter values that maximize the data fit.

\vspace{-6pt}
\subsection{Model training}
\vspace{-4pt}
To achieve co-estimation efficiently, the $T$-step ahead prediction cost on multiple overlapping subsections of the training data, used in the SUBNET method \citep{beintema_deep_2023}, is chosen to estimate the augmented model. In this way, model estimation using data sequence $\mathcal{D}_N=\{\left(y_i,u_i\right)\}_{i=1}^N$ acquired from \eqref{eq:DT-SS}, corresponds to an optimization problem with objective function $V^\mathrm{(sec)}_{\mathcal{D}_N}$, 
as
\begin{subequations}\label{eq:costfun}
\begin{align}
    V_{\mathcal{D}_N}^{(\mathrm{sec})}(\eta, \theta)&=\dfrac{1}{N_\mathrm{sec}} \sum_{k\in\mathcal{I}_{N_\mathrm{sec}}} \dfrac{1}{T} \sum_{l=0}^{T-1} \norm{y_{k+l} - \hat{y}_{k+l\vert k}}_2^2,\label{eq:V_sec}\\
        \hat{x}_{k+l+1\vert k} &= f_\theta(\hat{x}_{k+l\vert k}, u_{k+l}) + f^\mathrm{ANN}_\eta(\hat{x}_{k+l\vert k}, u_{k+l}),\\
        \hat{y}_{k+l\vert k} &= h_\theta(\hat{x}_{k+l\vert k}, u_{k+l})\\
        \hat{x}_{k\vert k} &= x_k,\label{eq:cost_fun_state_meas}
\end{align}
\end{subequations}
where the starting times of the sections collected in $\mathcal{I}_{N_\mathrm{sec}}$ are randomly selected from $\{1,\ldots, N - T + 1\} $, $N_\mathrm{sec} = | \mathcal{I}_{N_\mathrm{sec}}|$.
Here, $\hat{x}_{k+l\vert k}$ denotes the simulated state at $k+l$, starting from the measured initial state $\hat{x}_{k\vert k}=x_k$ at time  $k$. Training with the $T$-step ahead prediction cost brings two main benefits compared to traditional simulation error-based training: (\textit{i}) forward simulation on the subsections can be calculated in parallel, therefore requiring less computational load, and allowing the use of batch optimization algorithms; (\textit{ii}) using overlapping sections increases cost function smoothness \citep{ribeiro_smoothness_2020}, which further enhances the efficiency of (stochastic) gradient-based optimization methods.

Cost-function \eqref{eq:costfun} uses the initial state of each subsection for calculating the loss value, as visible in \eqref{eq:cost_fun_state_meas}. When full-state measurements are not applicable, i.e, when $y_k \neq x_k+e_k$, an encoder network is used to estimate the initial states from past input and output values, similarly as in \cite{beintema_deep_2023, hoekstra_learning-based_2024}.

\vspace{-8pt}
\subsection{Normalization and initialization scheme}\vspace{-4pt}
To efficiently train artificial neural networks, {\em input-output} (IO) normalization is essential to avoid exploding and vanishing gradients, which would potentially cause worse estimation results \citep{lecun_efficient_1998}.
A transformation for data normalization is defined according to \cite{schoukens_initialization_2020}:
\begin{align}
    T_u &= \mathrm{diag}\left(1/\sigma_{u,i}\right),\label{eq:Tu}\\
    T_x &= \mathrm{diag}\left(1/\sigma_{x,i}\right),\label{eq:Tx}
\end{align}
where $\sigma$ notes the standard deviation, and the transformation matrix $T_u\in\mathbb{R}^{n_u\times n_u}$ is a diagonal matrix containing the inverse of the standard deviation of $u_k$ for each channel $i$. The transformation matrix $T_x$ can be defined similarly. The standard deviations are calculated based on the training data, or, in case when state measurements are not available, transformation matrices are obtained by forward simulating the FP model on the training data.

Based on \eqref{eq:Tu} and \eqref{eq:Tx}, the state transition of the augmented model can be expressed as
\begin{equation}
    \hat{x}_{k+1} = f_\theta(\hat{x}_k,u_k) + T_x^{-1} f^\mathrm{ANN}_\eta(T_x \hat{x}_k, T_u u_k).
\end{equation}

Besides IO normalization, a reliable parameter initialization scheme is also important for efficient model training to improve accuracy and potentially decrease convergence time. A similar initialization method is applied as in \cite{ramkannan_initialization_2023}. This guarantees that the augmented state transition function behaves like $f_{\theta_0}$ at the beginning of training. 
A fully random initialization can increase convergence time and even cause stability loss at the beginning of the optimization. To prevent this, the weight and bias values of the last layer in the ANN\footnotemark{} are initialized as zero matrices, while the rest of the parameters are initialized randomly, e.g., by the Xavier approach \citep{glorot_understanding_2010}.
\footnotetext{Consider a simple feedforward neural network as
    $f^\mathrm{ANN}_\eta (\hat{x}_k,u_k)=W_{L+1} \sigma (W_L \dots \sigma(W_0 \left[ \hat{x}_k^\top\, u_k^\top \right]^\top + b_0) + b_L) + b_{L+1},$
where $\sigma(\cdot)$ is an element-wise (nonlinear) activation function, $W_i$ represents the weight values, $b_i$ notes the biases, and $L$ notes the number of hidden layers.}
\vspace{-8pt}
\section{Orthogonalization-based regularization}\label{sec:orthogonalization}
\vspace{-4pt}
Assuming that system \eqref{eq:DT-SS} is part of the model set spanned by the parametrization of \eqref{eqs:additive_struct}, there should exist a choice of parameter pairs $\left(\theta,\, \eta\right)$ for the additive model augmentation structure that realizes exactly the dynamical relations of the data-generating system. 
However, since $f^\mathrm{ANN}_\eta$ and $f_\theta$ are connected in parallel, i.e., additively, the optimal $\theta_\ast,\,\eta_\ast$ parameters that minimize \eqref{eq:costfun} are not unique, i.e.,
different $\theta,\,\eta$ parameters can result in the same input-state behavior. This is generally called non-identifiability. For ease of understanding, Example~\ref{example:non_uniquiness} illustrates this non-uniqueness.
\vspace{-4pt}
\begin{example}\label{example:non_uniquiness}
    Consider an FP model with a state-transition function of $f_\theta(\hat{x},u)=\theta \hat{x}$, and $f^\mathrm{ANN}_\eta$ as an ANN. By the universal approximation properties, there exist choices of weights and biases such that $f^\mathrm{ANN}_\eta(\hat{x},u) \approx W_{L+1} \hat{x}$. Consequently, for a true state-transition map $f(x,u)=\theta_\ast x$, any $\left(\theta,\,W_{L+1}\right)$ pair that satisfies $W_{L+1} + \theta = \theta_\ast$ is a global minimizer of \eqref{eq:costfun}.
\end{example}
\vspace{-6pt}

A straightforward approach would be to define a $\Theta_\ast\subset\mathbb{R}^{n_\theta+n_\eta}$ set that collects all parameter pairs that achieve an equivalent system representation with the data-generating system. Then, during model training, the aim would be to find any $\left(\theta_\ast,\, \eta_\ast\right) \in \Theta_\ast$. 
However, if the parameterization is non-identifiable, certain gradient directions become indistinguishable from one another, increasing the risk of convergence to local minima. While stochastic gradient descent-based optimizers inherently provide some degree of implicit regularization that helps to mitigate this issue \citep{zhang_understanding_2017}, another critical concern arises: the physical parameters may converge to unrealistic values due to the non-uniqueness of $\theta_\ast$. As a result, the estimated FP model cannot reliably serve as a standalone physical model of the system. 
Additionally, physics-based models are often capable of extrapolating accurately beyond the range of observed data, while ANNs extrapolate poorly outside of the training data set. Hence, the augmented model can lose this advantageous extrapolation capability due to the indistinguishability of the model contributions in the cost during training. 

We aim to address the above-mentioned problems 
by adding a new term to the cost function based on \cite{kon_physics-guided_2022} that forces $f^\mathrm{ANN}_\eta$ to prioritize the learning of the unknown dynamics. It is achieved by regularizing the output of $f^\mathrm{ANN}_\eta$ in the subspace where $f_\theta$ can generate outputs. The original approach has been derived for feedforward control, but it can be easily modified for system identification problems if the FP model can be represented as being {\em linear in the parameters}:
\begin{equation}
    f_\theta(\hat{x}_k, u_k)=\phi(\hat{x}_k,u_k) \theta.
\end{equation}

By taking a further assumption of full state-measurement, i.e., $y_k=x_k+e_k$, and fixing the learning component to zero, due to the linear parameterization, fitting of the FP model parameters becomes a {\em linear regression} (LR) problem:
\begin{equation}\label{eq:F(X,U)}\vspace{-4pt}
    \underbrace{\begin{bmatrix}
        x_1\\x_2\\\vdots\\x_N
    \end{bmatrix}}_{X^+} = \underbrace{\begin{bmatrix}
        \phi(x_0, u_0)\\
        \phi(x_1, u_1)\\
        \vdots\\
        \phi(x_{N-1}, u_{N-1})
    \end{bmatrix}}_{\Phi(X,U)} \theta + \underbrace{\begin{bmatrix}
        e_1\\ e_2\\ \vdots\\ e_N
    \end{bmatrix}}_E,\vspace{-4pt}
\end{equation}
where $X=\begin{bmatrix}
    x_0^\top & x_1^\top & \cdots & x_{N-1}^\top
\end{bmatrix}^\top \in \mathbb{R}^{Nn_x}$, and $U$ is defined similarly. The $E$ matrix collects the residuals of the FP state transition. Given dataset $\mathcal{D}_N$, an explicit base for the output space of $f_\theta$ can be calculated by taking the reduced {\em singular value decomposition} (SVD) of $\Phi(X,U)\in\mathbb{R}^{Nn_x\times n_\theta}$, as\vspace{-4pt}
\begin{equation}
    \Phi(X,U)=
        Q_{X,U} \Sigma_{X,U} V^\top_{X,U},
\end{equation}
in which we assume $Nn_x>n_\theta$ such that $Q_{X,U}\in\mathbb{R}^{Nn_x\times n_\theta}$, $\Sigma_{X,U}\in\mathbb{R}^{n_\theta \times n_\theta}$, and $V_{X,U}\in\mathbb{R}^{n_\theta\times n_\theta}$. Thus, \eqref{eq:F(X,U)} can be reformulated as\vspace{-4pt}
\begin{equation}
    X^+=Q_{X,U} \Sigma_{X, U} V^\top_{X,U} \theta + E.\label{eq:decomposed_state}\vspace{-4pt}
\end{equation}

The columns of $Q_{X,U}$ form a basis for the output space of $f_\theta$ for the particular state and input values in $X,\,U$. The projection matrix onto the subspace spanned by these basis vectors can be expressed as\vspace{-4pt}
\begin{equation}
    \Pi_{X,U}=Q_{X,U} Q^\top_{X,U}.\label{eq:Pi}\vspace{-4pt}
\end{equation}

Thus, the component of $f^\mathrm{ANN}_\eta$ that lies in the subspace spanned by $Q_{X,U}$ can be given by $\Pi_{X,U} f^\mathrm{ANN}_\eta(X,U)$, where $f^\mathrm{ANN}_\eta(X,U)=\left[
    f^\mathrm{ANN}_\eta(x_0,u_0)^\top\,\cdots\, f^\mathrm{ANN}_\eta(x_{N-1},u_{N-1})^\top
\right]^\top \in \mathbb{R}^{N n_x}$. The aim is to penalize such $\eta$ parameters that result in significant contributions in this subspace, by adding $\norm{\Pi_{X,U} f^\mathrm{ANN}_\eta(X,U)}_2^2$ to \eqref{eq:V_sec}, as an orthogonality-promoting term. Thus, it can be interpreted as a targeted $\ell_2$ regularization of the ANN that only penalizes the directions of $\eta$ which generate output in the subspace of the model $f_\theta$. The modified cost function is given as
\begin{equation}
    V_{\mathcal{D}_N}^{(\mathrm{orth})}(\eta, \theta)=V_{\mathcal{D}_N}^{(\mathrm{sec})}(\eta, \theta) + \beta \norm{\Pi_{X,U} f^\mathrm{ANN}_\eta(X,U)}_2^2,\label{eq:orth_cost}
\end{equation}
where $\beta\in\mathbb{R}_{\geq 0}$ is the orthogonalization coefficient. The value of $\beta$ can be interpreted as a trade-off between orthogonality and performance. With large $\beta$ values, $f^\mathrm{ANN}_\eta$ is restricted too much, and the optimizer focuses more on promoting orthogonality between the FP model and the ANN part, rather than producing an accurate model. On the other hand, too small $\beta$ values can lead to a near unregularized situation by diminishing the penalization of subspace contributions.
\vspace{-6pt}
\begin{remark}
In practice, often only a small amount of orthogonal regularization is needed to achieve the desired complementarity. A wide range of $\beta$ values can be chosen, usually spanning several orders of magnitude. Later we will illustrate this in our example (in Figure~\ref{fig:beta_tuning}). A similar observation has been made in \cite{kon_learning_2023}.
\end{remark}
\vspace{-6pt}

Since $N n_x \gg n_\theta$, matrix multiplication with $\Pi_{X,U}\in\mathbb{R}^{N n_x\times N n_x}$ is computationally demanding in $\Pi_{X,U}f^\mathrm{ANN}_\eta(X,U)$. However, since $\Pi_{X,U} = Q_{X,U} Q_{X,U}^\top$, and $Q_{X,U}$ is orthonormal, it holds that
\begin{equation}
    \norm{\Pi_{X,U}f^\mathrm{ANN}_\eta(X,U)}_2^2 = \norm{Q_{X,U}^\top f^\mathrm{ANN}_\eta(X,U)}_2^2.\label{eq:U_top}
\end{equation}
Calculating \eqref{eq:U_top} is less intensive, as it is composed of $n_\theta$ numbers of dot product calculation between $f^\mathrm{ANN}_\eta$ and vectors of length $N n_x$. 
Another advantage of the decomposition shown in \eqref{eq:decomposed_state} is that the projection matrix $\Pi_{X,U}$ is independent of the physical parameters since $\Phi$ depends only on the state and input values. This practically means that the SVD can be calculated before the optimization process begins, thus lowering the computational demand of the method. When full-state measurements are not available, an approximate $\hat{X}$ data set can be constructed by forward simulating the FP model on the training data, and using $\hat{X}$ to compute $\Pi_{\hat{X},U}$.
\vspace{-6pt}
\begin{remark}
The current model estimates of $\hat{X}$ can be also used to update the projection matrix at each iteration step. This modification increases computational demand, as the SVD needs to be recomputed at the start of each epoch. However, the benefit of it is that the regularization term is applied for the accurate states, potentially leading to faster convergence and better model accuracy.
\end{remark}
\vspace{-20pt}
\section{Generalization for nonlinear first-principle models}\label{sec:orth_generalization}
\vspace{-6pt}
A serious limitation of the orthogonalization-based method derived in Sect.~\ref{sec:orthogonalization} is that it can only be applied to FP models that are {\em linear-in-the-parameters}. To extend the method to the nonlinear case, the Taylor series expansion of the FP state transition function is taken w.r.t. the physical parameters:
\begin{equation}
    f_\theta(\hat{x}_k,u_k) \approx f_\theta(\hat{x}_k,u_k) \bigr\vert_{\theta=\bar{\theta}} + \underbrace{\pdv{f_\theta(\hat{x}_k,u_k)}{\theta} \Bigr\vert_{\theta=\bar{\theta}}}_{\Phi_{\bar{\theta}}(\hat{x}_k,u_k)} \left(\theta - \bar{\theta}\right),\label{eq:f_taylor}\vspace{-4pt}
\end{equation}
where $\bar{\theta}$ is the linearization point, and $\partial f_\theta(\hat{x}_k,u_k)/\partial \theta \in \mathbb{R}^{n_x \times n_\theta}$ is the Jacobian matrix of $f_\theta(\hat{x}_k,u_k)$ with respect to the $\theta$ parameters.
Next, expanding \eqref{eq:f_taylor}, we arrive to the expression of\vspace{-4pt}
\begin{equation}
    f_\theta(\hat{x}_k,u_k) \approx \Phi_{\bar{\theta}}(\hat{x}_k,u_k) \theta + f_{\bar{\theta}} (\hat{x}_k,u_k) -  \Phi_{\bar{\theta}}(\hat{x}_k,u_k) \bar{\theta},\vspace{-4pt}
\end{equation}
where the first term is linear in $\theta$, while the second and third terms are a $\theta$-independent offset given by the linearization point $\bar{\theta}$. Similar to \eqref{eq:F(X,U)}, by vectorizing the approximate FP model responses,
\begin{equation}
        \underbrace{\begin{bmatrix}
        x_1\\x_2\\\vdots\\x_N
    \end{bmatrix}}_{X^+} = \underbrace{\begin{bmatrix}
        \Phi_{\bar{\theta}}(x_0,u_0)\\
        \Phi_{\bar{\theta}}(x_1,u_1)\\
        \vdots\\
        \Phi_{\bar{\theta}}(x_{N-1},u_{N-1})
    \end{bmatrix}}_{\Phi_{\bar{\theta}}(X,U)} \theta + \underbrace{\begin{bmatrix}
        f_{\bar{\theta}}(x_0,u_0) - \Phi_{\bar{\theta}}(x_0,u_0) \bar{\theta}\\
        f_{\bar{\theta}}(x_1,u_1) - \Phi_{\bar{\theta}}(x_1,u_1) \bar{\theta}\\
        \vdots\\
        f_{\bar{\theta}}(x_{N-1},u_{N-1}) - \Phi_{\bar{\theta}}(x_{N-1},u_{N-1}) \bar{\theta}
    \end{bmatrix}}_{\Gamma_{\bar{\theta}}(X,U)} + \underbrace{\begin{bmatrix}
        e_1\\ e_2\\ \vdots\\ e_N
    \end{bmatrix}}_E.\label{eq:orth_general}\vspace{-4pt}
\end{equation}
To consider all components of the baseline model in promoting orthogonality, and to enable adjustments to the linearization point during model learning, an extended parameter vector is introduced as $\tilde{\theta}\in \mathbb{R}^{n_\theta+1}$. Then, the two terms in \eqref{eq:orth_general} can be combined:\vspace{-4pt}
\begin{equation}
    X^+ = \underbrace{\begin{bmatrix}
        \Phi_{\bar{\theta}}(X,U) & \Gamma_{\bar{\theta}}(X,U)
    \end{bmatrix}}_{\tilde{\Phi}_{\bar{\theta}}(X,U)} \underbrace{\begin{bmatrix}
        \theta\\ 1
    \end{bmatrix}}_{\tilde{\theta}} + E.\label{eq:orth_general_final_form}\vspace{-4pt}
\end{equation}

It is evident that \eqref{eq:orth_general_final_form} has a structure similar to \eqref{eq:F(X,U)}, hence the same orthogonalization-based penalization term can be derived for it, but instead of taking the reduced SVD of $\Phi(X,U)$, now it should be calculated for $\tilde{\Phi}_{\bar{\theta}}(X,U)$. The projection matrix $\tilde{\Pi}_{X,U}$ can then be calculated as in \eqref{eq:Pi}.
Similarly to \eqref{eq:orth_cost}, the orthogonality-promoting term is added to the cost function, as\vspace{-4pt}
\begin{equation}
    V_{\mathcal{D}_N}^{(\mathrm{orth})}(\eta, \theta)=V_{\mathcal{D}_N}^{(\mathrm{sec})}(\eta, \theta) + \beta \norm{\tilde{\Pi}_{X,U} f^\mathrm{ANN}_\eta(X,U)}_2^2.\vspace{-4pt}
\end{equation}

The linearization point $\bar{\theta}$ can be updated at each iteration using the current estimate of $\theta$, but this requires recomputing the SVD and projection matrix every time the cost function is evaluated, increasing computational demand. Alternatively, $\bar{\theta}$ can be approximated by the nominal parameter values $\theta_0$, allowing the projection matrix to be precomputed at the start of training, which significantly reduces the computational cost of the regularization method. 
As $\theta_0$ is derived from physical principles and typically aligns well with the actual system properties, the approximation remains valid and the basis for the response of the physical model is accurately maintained. 
\vspace{-12pt}
\section{Identification study}\label{sec:f1tenth_ident}
\vspace{-6pt}
To demonstrate the advantages of the proposed orthogonal projection-based regularization for learning-based model augmentation, we aim to identify the dynamics of a small-scale electric vehicle (F1Tenth \citep{agnihotri_teaching_2020}), by augmenting with an ANN component a simplified mechanical model that describes the main characteristics of the vehicle\footnote{Code, data available: \url{https://github.com/AIMotionLab-SZTAKI/orthogonal-augmentation}}.
%

\vspace{-2pt}
\begin{figure}
    \centering
    \begin{minipage}{0.45\textwidth}
        \centering
        \includegraphics[scale=0.7]{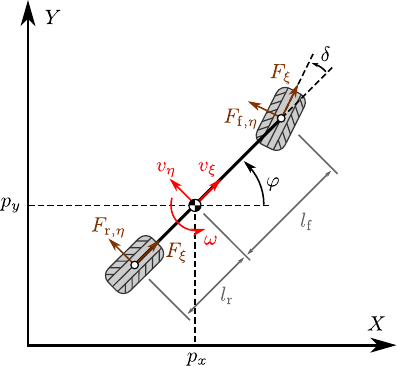}\vspace{-6pt}
        \caption{The single track model.}
        \label{fig:single_track}
    \end{minipage}
    \hfill
    \begin{minipage}{0.45\textwidth}
        \centering
        \includegraphics[width=0.78\linewidth]{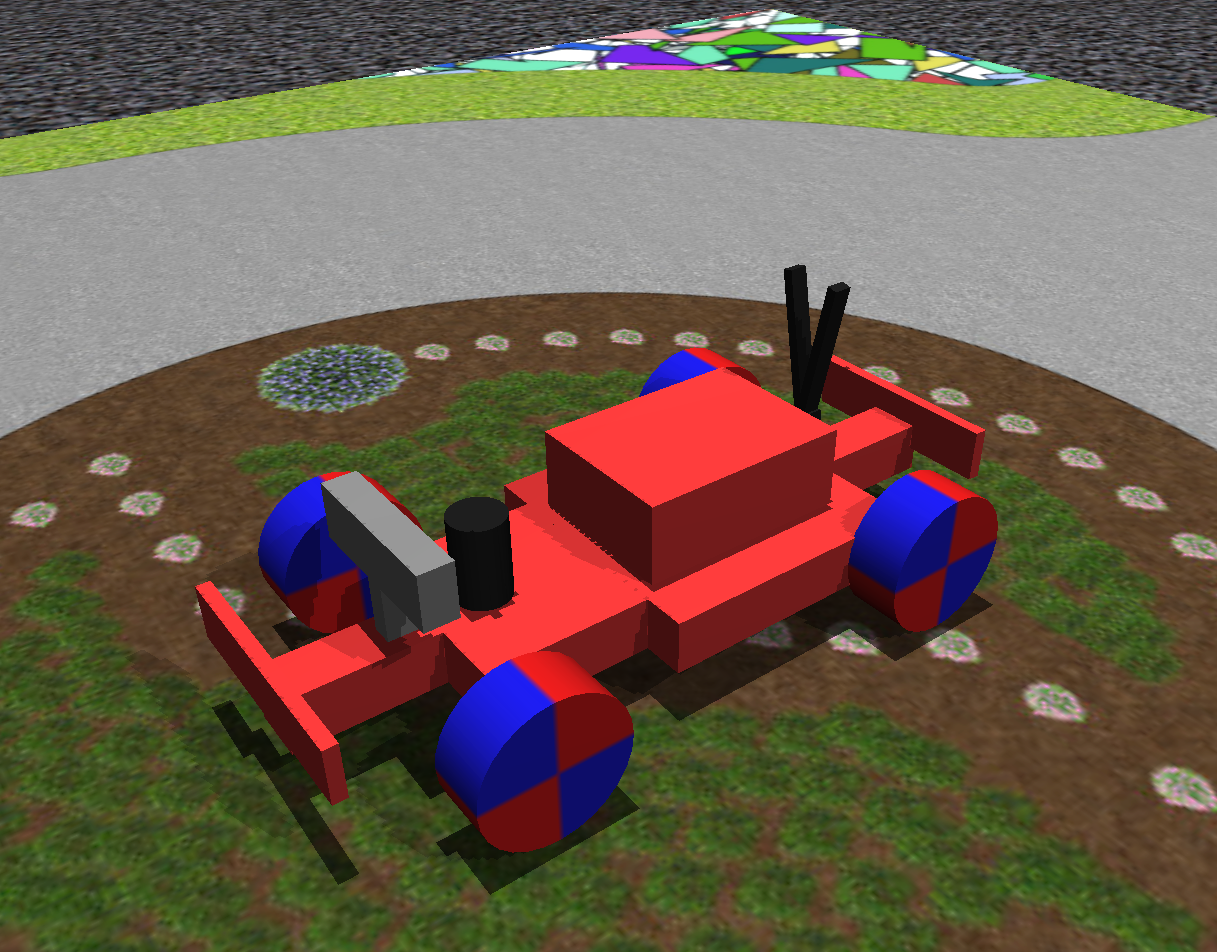}
        \caption{Digital twin of the F1tenth vehicle.}
        \label{fig:digital_twin}
    \end{minipage}
    \vspace{-12pt}
\end{figure}

\vspace{-8pt}
\subsection{First-principle model of the F1Tenth vehicle}\vspace{-6pt}
To develop an approximate baseline model of the F1tenth platform, the so-called single-track model has been used \citep{paden_survey_2016}. The model is illustrated in Figure~\ref{fig:single_track}, and expressed as\vspace{-2pt}
\begin{subequations}\label{eq:car_EoM}
\begin{align}   
    \dot{p}_x&=v_\xi \cos \varphi - v_\eta \sin \varphi,\label{eq:car_x} & \dot{v}_\xi&=\dfrac{1}{m} \left(F_\xi+F_\xi \cos\delta - F_\mathrm{f,\eta} \sin\delta + m v_\eta \omega\right),\\
    \dot{p}_y&=v_\xi \sin \varphi + v_\eta \cos \varphi,\label{eq:car_y} & \dot{v}_\eta&=\dfrac{1}{m} \left(F_\mathrm{r,\eta}+F_\xi \sin\delta - F_\mathrm{f,\eta} \cos\delta - m v_\xi \omega \right),\\
    \dot{\varphi}&=\omega,\label{eq:car_phi} & \dot{\omega}&=\dfrac{1}{J_\mathrm{z}} \left(F_\mathrm{f,\eta} l_\mathrm{f} \cos\delta +F_\xi l_\mathrm{f} \sin\delta - F_\xi l_\mathrm{r}\right),\vspace{-6pt}
\end{align}
\end{subequations}
where $\left(p_x,p_y\right)$ is the position of the \emph{center-of-gravity} (CoG) in the $\left(X,Y\right)$ plane, $\varphi$ is the orientation of the vehicle, which is measured from the $X$ axis. The variables $v_\xi$ and $v_\eta$ denote the longitudinal and lateral velocities of the vehicle, respectively, while $\omega$ is the yaw rate. Furthermore, $l_\mathrm{r}$ and $l_\mathrm{f}$ are the distances of the rear and front axis from the CoG, $\delta$ is the steering angle, $m$ is the mass of the vehicle, and $J_z$ is the inertia along the vertical axis.

The longitudinal tire force component $F_\xi$ is expressed with an empirical drivetrain model. For more details, refer to \cite{floch_gaussian-process-based_2024}. Moreover, the linearized Magic Formula \citep{pacejka_chapter_2012} is utilized to model the lateral tire forces. Thus, the tire force components are expressed as\vspace{-4pt}
\begin{equation}
    F_\xi = C_\mathrm{m1}d - C_\mathrm{m2} v_\xi - \mathrm{sign}\left(v_\xi\right) C_\mathrm{m3},\quad F_\mathrm{r,\eta}=C_\mathrm{r} \frac{-v_\eta+l_\mathrm{r}\omega}{v_\xi},\quad F_\mathrm{f,\eta}=C_\mathrm{f} \left(\delta - \frac{v_\eta+l_\mathrm{f}\omega}{v_\xi}\right),
\end{equation}
where $C_\mathrm{m1}$, $C_\mathrm{m2}$, and $C_\mathrm{m3}$ are the drivetrain constants, $d$ is the motor PWM percentage, $F_\mathrm{r,\eta}$ and $F_\mathrm{f,\eta}$ are the rear and front lateral tire forces, while $C_\mathrm{r}$ and $C_\mathrm{f}$ are the rear and front cornering stiffness values, respectively. The applied tire models (particularly the empirical drivetrain model) are highly approximative and are the primary sources of inaccuracy in the FP model.

The state of the baseline model is $x=\begin{bmatrix}
    p_x & p_y & \varphi & v_\xi & v_\eta & \omega
\end{bmatrix}^\top$. The control inputs are $\delta$ and $d$. As discussed in Sect.~\ref{sec:additive_augm}, we have assumed that the FP model is given in DT form, thus \eqref{eq:car_EoM} is discretized with the forward Euler scheme. The parameters of the baseline model are\vspace{-4pt}
\begin{equation}\label{eq:f1tenth_theta}
    \theta=\begin{bmatrix}
        m & J_z & l_\mathrm{r} & l_\mathrm{f} & C_\mathrm{m1} & C_\mathrm{m2} & C_\mathrm{m3} & C_\mathrm{r} & C_\mathrm{f}
    \end{bmatrix}^\top,
    \vspace{-4pt}
\end{equation}
and their nominal values have been determined in \cite{floch_model-based_2022}. However, to achieve accurate representation of the true dynamics, $\theta$ in \eqref{eq:f1tenth_theta} is tuned jointly with the ANN parameters.

As one might see, in \eqref{eq:car_EoM}, the differential equations for the position and orientation values contain simple kinematic relations and can be separated from the rest of the equations. As proposed by \cite{szecsi_deep_2024}, to simplify the neural network structure, we only augment the velocity states of the system, then $p_x$, $p_y$, and $\varphi$ can be determined by numerical integration.

\vspace{-12pt}
\subsection{Data acquisition}\vspace{-6pt}
To generate data for training, a high-fidelity multi-body simulator, the MuJoCo physics engine
is utilized \citep{todorov_mujoco_2012}. In this simulator, the digital twin model of the car was assembled using parameters that were identified based on measurements with the real car. Figure~\ref{fig:digital_twin} shows the digital twin model. Multiple experiments were performed in the simulator to acquire sufficient data. The MuJoCo model operates with motion states in terms of joint positions and velocities, utilizing complex contact models and friction characteristics. The logged signals have been determined based on the states of the FP model. All states are directly measured, as it would be for the real F1Tenth. Two different trajectories (a lemniscate and a circular path), with 12 different velocity references have been utilized to achieve robust identification results. With a sampling frequency of $f_\mathrm{s}=40~\mathrm{Hz}$, altogether $N=15985$ data points have been recorded. Half of the measurements were separated for model training, while the other half were used for testing. Trajectories with alternating reference velocities have been included in both. Lastly,
20\% of each training trajectory have been randomly selected to form a validation data set, achieving an 80\%-20\% ratio of the training and validation data points. The gathered data does not contain any noise, but to test the methodology with noise levels that are typical for this application type, i.i.d. white Gaussian noise is added to the output signals in the training and validation data sets to reach {\em Signal-to-noise ratio}\footnote{$\mathrm{SNR_{dB}}=10\log_{10}\left(P_\mathrm{s}/P_\mathrm{n}\right)$, where $P_\mathrm{s}$ and $P_\mathrm{n}$ are the sample mean signal power and noise power, respectively.} (SNR) values of 30~dB and 25~dB. To ensure a clear comparison of the resulting models, the test data is kept noise-free.
%
%

\vspace{-10pt}
\subsection{Model training and hyperparameter selection}\vspace{-6pt}
All models have been trained with the Adam optimizer \citep{kingma_adam_2015}, applying a learning rate of $10^{-3}$ and a batch size of 256. For the truncation length, $T=15$ has been used, corresponding to roughly 1.5 times the largest characteristic time scale of the data-generating system. The hyperparameters of $f^\mathrm{ANN}_\eta$ have been determined empirically: 2 hidden layers with 64 nodes per layer have been applied, using the {\em hyperbolic tangent} (tanh) activation function. 
The regularization coefficient $\beta$ was also tuned empirically, as illustrated in Figure~\ref{fig:beta_tuning}, with $\beta=10^{-7}$ selected for the noiseless case. For scenarios with output noise, $\beta$ was chosen using a similar approach. Based on our experience, values of $\beta$ in the range $\left[10^{-7},\,10^{-5}\right]$ yielded the best results for the considered modeling task.

\vspace{-10pt}
\subsection{Results}\vspace{-6pt}
%
\begin{figure}
  \begin{minipage}[c]{.38\linewidth}
    \centering
    \includegraphics[width=\linewidth]{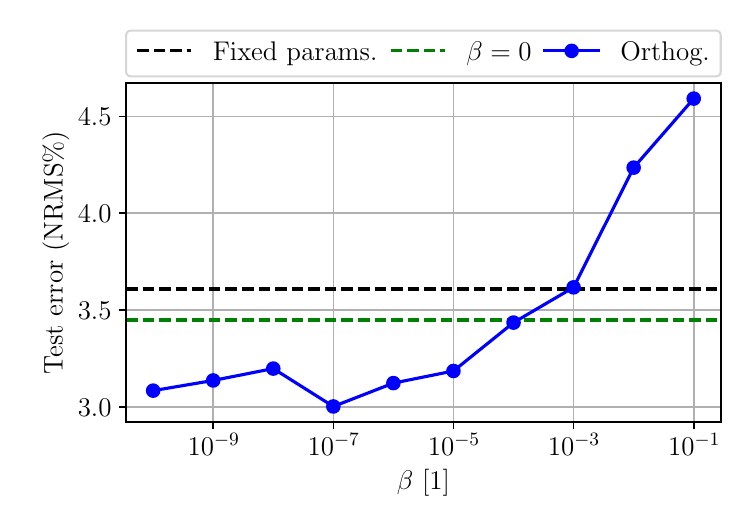}\vspace{-14pt}
    \captionof{figure}{Illustrating the empirical process of tuning $\beta$ (with no  noise).}
    \label{fig:beta_tuning}
  \end{minipage}
  \hfill
  \begin{minipage}[c]{.6\linewidth}
    \footnotesize
    \centering
    \begin{tabular}{lccc}
        \hline
        Model & No noise & 30 dB & 25 dB\\
        \hline
        Initial baseline model with $\theta_0$ & 37.91\% & 37.91\% & 37.91\%\\
        Baseline model with $\hat{\theta}$ (no reg.) & 118.01\% & 80.49\% & 68.18\%\\
        Baseline model with $\hat{\theta}$ (orth. reg.) & 36.07\% & 36.64\% & 30.69\%\\
        Augm. model (fixed baseline $\theta_0$) & 3.61\% & 4.18\% & 5.11\%\\
        Augm. model (co-estim., no reg.) & 3.45\% & 4.17\%  & 4.92\%\\
        \textbf{Augm. model (co-estim., orth. reg.)} & \textbf{3.00\%} & \textbf{3.84\%} & \textbf{4.76\%} \\
        Black-box \citep{beintema_deep_2023} & 2.12\% & 2.31\% & 3.06\%\\
        \hline
    \end{tabular}\vspace{-6pt}
    \captionof{table}{Test NRMS errors with various models. SNR values refer to the noise level in the training and validation sets.}
    \label{tab:results}
  \end{minipage}
  \vspace{-18pt}
\end{figure}
%
%
%
Model performance is evaluated using the {\em Normalized Root Mean Square} (NRMS) error, see, e.g., \cite{beintema_nonlinear_2021}, with results summarized in Table~\ref{tab:results}. As shown, augmenting the baseline model while keeping the physical parameters fixed leads to a significant improvement in accuracy compared to the standalone FP model. Jointly optimizing the physical and ANN parameters has further reduced the test error, while using the orthogonalization-based cost function led to additional improvements in model accuracy across all noise conditions. With increasing noise levels, the effect of orthogonal regularization on model accuracy is less noticeable compared to the noiseless scenario, but this is expected, as it is harder to separate the baseline model dynamics and the noise.

When the physical and learning-based parameters are co-estimated, we acquire a $(\hat{\theta},\,\hat{\eta})$ parameter pair as a result of the optimization. This allows for evaluating the FP model separately using the estimated physical parameters $\hat{\theta}$, providing insight into the interpretability of the augmented model. Table~\ref{tab:results} also includes these test results with the FP model, using $\hat{\theta}$. 
Without regularization, the physics-based component of the augmentation structure shows a notable drop in accuracy compared to the nominal baseline model, supporting our earlier statement that, in the absence of the proposed regularization, physical parameters may be tuned to unrealistic values, potentially undermining the interpretability of the augmented model. 
In contrast, applying orthogonal regularization not only enhances the overall accuracy of the augmented model but also improves the performance of the physics-based part relative to the nominal FP model. 
This demonstrates that the proposed method effectively promotes the desired complementarity between the baseline and learning components. 

Comparing the results to a state-of-the-art black-box method, it is visible that even with the orthogonal projection-based approach, black-box models have performed slightly better.
However, as we have demonstrated, model augmentation generates physically interpretable models in contrast to black-box approaches. Consequently, a slight reduction in model accuracy is an acceptable trade-off for the added interpretability. 

\vspace{-16pt}
\section{Conclusion}\label{sec:conclusion}\vspace{-8pt}
This paper introduced an orthogonal projection-based regularization method that extends the approach in \cite{kon_physics-guided_2022}, originally limited to linear-in-the-parameter FP models, to support models that are nonlinear in the parameters and improve the efficiency of estimating additive model augmentation structures. 
The proposed method improved model performance across various output noise levels. 
Although the impact of online recalculation of the projection matrix has not been evaluated, its potential benefits, despite the increased computational cost, will be explored in future studies. 
Further theoretical analysis and application in other model augmentation settings will be addressed in future research.


\acks{{\small This project has received funding from the European Defence Fund programme under grant agreement number No 101103386 and has also been supported by the Air Force Office of Scientific Research under award number FA8655-23-1-7061 and by the EKÖP-24-2-BME-123 University Research Scholarship Programme of the Ministry for Culture and Innovation from the source of the National Research, Development, and Innovation Fund. Views and opinions expressed are however those of the authors only and do not necessarily reflect those of the European Union or the European Commission. Neither the European Union nor the granting authority can be held responsible for them. The study was also partly funded by the National Research, Development and Innovation Office in Hungary (RRF-2.3.1-21-2022-00006).}}

\bibliography{literature.bib}

\end{document}